\documentclass[letterpaper]{article} 
\usepackage{aaai25}  
\usepackage{times}  
\usepackage{helvet}  
\usepackage{courier}  
\usepackage[hyphens]{url}  
\usepackage{graphicx} 
\usepackage{natbib}  
\usepackage{caption} 
\usepackage{amssymb}
\usepackage{xcolor}
\usepackage{amsmath}
\usepackage{algorithm}
\usepackage{algorithmic}
\usepackage{multirow}
\usepackage{colortbl}
\usepackage{xcolor,soul}
 
\usepackage{newfloat}
\usepackage{listings}
\DeclareCaptionStyle{ruled}{labelfont=normalfont,labelsep=colon,strut=off} 
\floatstyle{ruled}
\newfloat{listing}{tb}{lst}{}
\floatname{listing}{Listing}
\title{CAPrompt: Cyclic Prompt Aggregation for Pre-Trained Model Based \\Class Incremental Learning}
\author{
    Qiwei Li,
    Jiahuan Zhou\thanks{Corresponding author}
}
\affiliations{
    Wangxuan Institute of Computer Technology, Peking University, Beijing 100871, China\\

    \{lqw, jiahuanzhou\}@pku.edu.cn
}

\usepackage{bibentry}

\begin{document}

\maketitle

\begin{abstract}
Recently, prompt tuning methods for pre-trained models have demonstrated promising performance in Class Incremental Learning (CIL). These methods typically involve learning task-specific prompts and predicting the task ID to select the appropriate prompts for inference. However, inaccurate task ID predictions can cause severe inconsistencies between the prompts used during training and inference, leading to knowledge forgetting and performance degradation. Additionally, existing prompt tuning methods rely solely on the pre-trained model to predict task IDs, without fully leveraging the knowledge embedded in the learned prompt parameters, resulting in inferior prediction performance. To address these issues, we propose a novel Cyclic Prompt Aggregation (CAPrompt) method that eliminates the dependency on task ID prediction by cyclically aggregating the knowledge from different prompts. Specifically, rather than predicting task IDs, we introduce an innovative prompt aggregation strategy during both training and inference to overcome prompt inconsistency by utilizing a weighted sum of different prompts. Thorough theoretical analysis demonstrates that under concave conditions, the aggregated prompt achieves lower error compared to selecting a single task-specific prompt. Consequently, we incorporate a concave constraint and a linear constraint to guide prompt learning, ensuring compliance with the concave condition requirement. Furthermore, to fully exploit the prompts and achieve more accurate prompt weights, we develop a cyclic weight prediction strategy. This strategy begins with equal weights for each task and automatically adjusts them to more appropriate values in a cyclical manner. Experiments on various datasets demonstrate that our proposed CAPrompt outperforms state-of-the-art methods by 2\%-3\%. Our code is available at https://github.com/zhoujiahuan1991/AAAI2025-CAPrompt.
\end{abstract}

%

\section{Introduction}

\begin{figure}[t]
    \centering
	\includegraphics[width=1\linewidth]{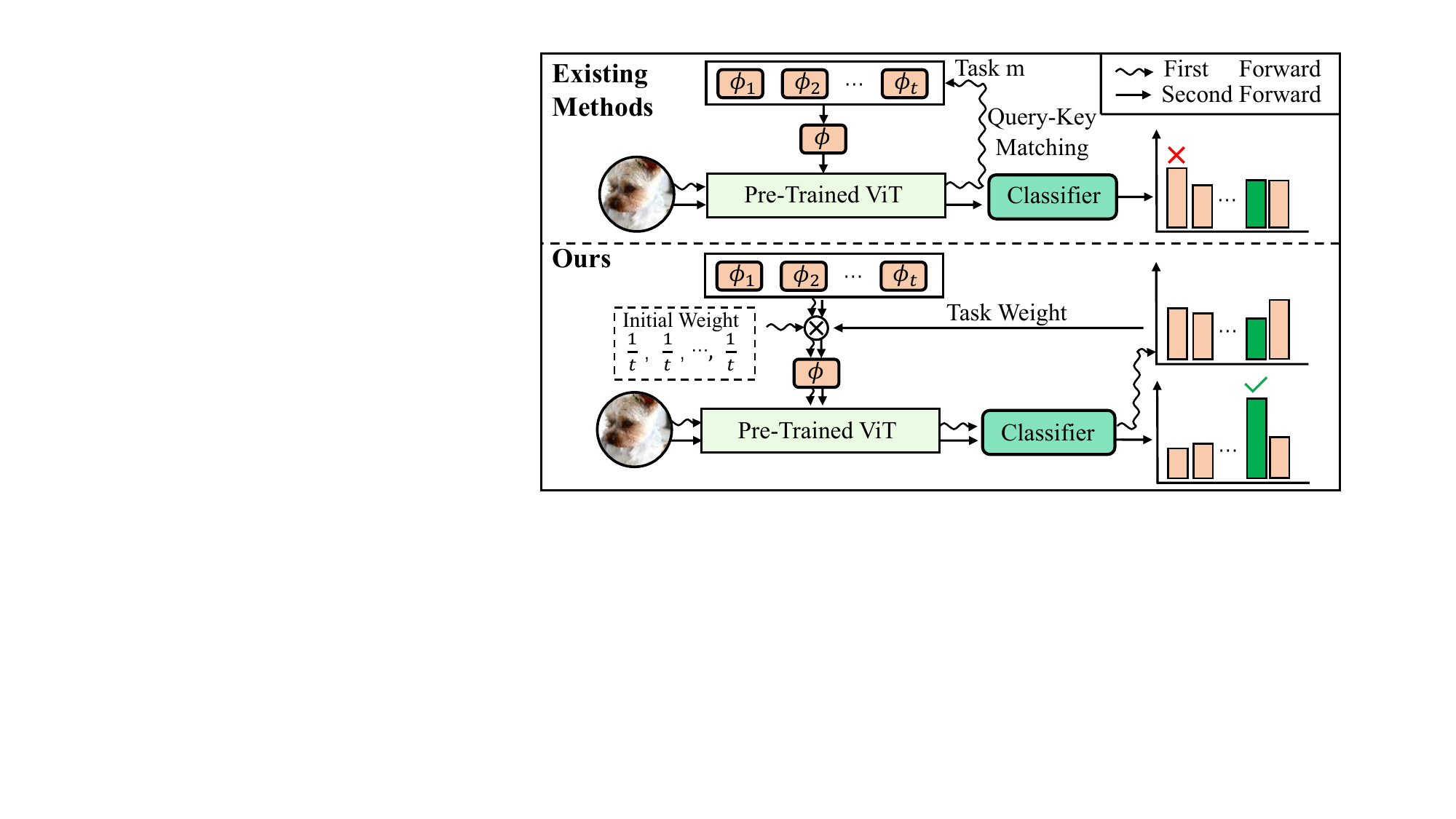}
        \caption{\label{fig:motivation1} Most existing prompt-based methods (Dualprompt) predict task ID during inference which may cause inconsistency between the prompts during training and inference. In contrast, we propose a prompt aggregation strategy to eliminate the requirement to predict task ID. Moreover, a cyclic prompt weight strategy is proposed to adjust the weights of different prompts.}
    
\end{figure}

In recent years, the field of computer vision has seen remarkable advancements, particularly in the domain of image classification~\cite{lecun1989handwritten,deng2009imagenet,dosovitskiy2020image}. However, traditional learning models typically assume that the entire training dataset is available from the onset~\cite{krizhevsky2012imagenet,he2016deep}. This assumption leads to the challenge when models are required to adapt to new data incrementally without forgetting previously learned information, a phenomenon commonly referred to as \textit{catastrophic forgetting}~\cite{french1999catastrophic}. \textit{Class Incremental Learning} (CIL) \cite{zhou2024class,wang2024comprehensive} aims to address this issue by enabling models to learn continuously from a stream of data, adapting to new tasks while retaining past knowledge.

The primary challenge of CIL is balancing the acquisition of new knowledge and the retention of previous knowledge \cite{zhou2024class}. A promising solution lies in leveraging pre-trained models, which are trained on large, diverse datasets to capture rich and generalized feature representations. Recent methods aim to utilize Parameter-Efficient Fine-Tuning (PEFT) techniques, such as prompt learning \cite{wang2022learning,wang2022dualprompt}, to acquire new knowledge while keeping the pre-trained model frozen, thereby mitigating forgetting. During training, these methods not only learn a set of task-specific prompt parameters but also feed the input sample into the pre-trained model without prompts to learn a set of keys for each task, which are used for task selection by query-key matching during inference. As demonstrated in Fig.\ref{fig:motivation1}, in the inference stage, due to the absence of task ID information, these methods need to infer the task ID based on the learned keys at first. In the subsequent inference stage, the predicted task ID is used to select the relevant task-specific prompt for the pre-trained model. However, the predicted task ID for certain samples may not match their ground truth ID, resulting in the adoption of prompts associated with incorrect tasks during inference. This inconsistency between the prompts used during training and inference inevitably leads to severe knowledge forgetting, thereby decreasing final performance \cite{gao2024consistent}. Furthermore, existing methods rely solely on the pre-trained model to predict task ID without fully leveraging the knowledge embedded in the learned prompt parameters, exacerbating the degradation in performance.

To address these challenges, we propose a novel \textbf{C}yclic \textbf{Prompt} \textbf{A}ggregation (\textbf{CAPrompt}) method for CIL with pre-trained models. As illustrated in Fig.\ref{fig:motivation1}, instead of predicting task IDs to select one task-specific prompt, we estimate the probability of each class and aggregate these probabilities into the probability of each task. These task probabilities are then used as weights for prompts at different tasks to generate the aggregated prompt. Unlike a few existing methods that employ similar weighted sums of prompts \cite{smith2023coda,roy2024convolutional}, we provide a theoretical justification that the prediction error of the aggregated prompt is lower than using one task-specific prompt parameter in a single stage, under the condition that the network prediction is concave with respect to the prompt parameters for a given image. To ensure the network satisfies this concave condition, we introduce a concave constraint and a linear constraint. Furthermore, to fully utilize the knowledge embedded in the prompts to guide the pre-trained models, we develop a cyclic prompt weight prediction strategy. Instead of merely predicting the prompt weight without the guidance of prompts, we initialize with equal weights for each task and cyclically adjust the prompt weights to more appropriate values. Additionally, beyond the conventional two-stage paradigm, our method can be performed cyclically multiple times, further boosting performance.

In summary, the contributions of this paper are four-fold: (1) A new Prompt Aggregation strategy is proposed for both training and inference, eliminating the need for task ID prediction thus mitigating the inconsistencies of prompts caused by task prediction errors. (2) Comprehensive theoretical analysis is provided to demonstrate that our aggregated prompt achieves lower prediction error compared to using a single task-specific prompt under the concave condition. The concave and linear constraints are proposed to facilitate the model to satisfy this condition. (3) A cyclic prompt weight prediction strategy is proposed to cyclically adjust the weights of prompts to more accurate ones, further improving performance. (4) Extensive experiments on various benchmarks demonstrate that our CAPrompt outperforms state-of-the-art approaches by 2\%-3\%.

\section{Related Work}
\subsection{Class Incremental Learning}
Current Class Incremental Learning methods can be broadly categorized into \textit{rehearsal-based}, \textit{regularization-based}, and \textit{architecture-based} methods. Rehearsal-based approaches~\cite{prabhu2020gdumb,liu2021rmm,luo2023class} selected and stored representative samples from earlier classes to explicitly preserve knowledge. During the training of later classes, they replayed these stored samples to mitigate forgetting. Moreover, regularization-based methods~\cite{kirkpatrick2017overcoming,li2017learning,smith2021always,li2024exemplar,li2024progressive,li2024fcs,xu2024lstkc} stabilized model parameters and feature adjustments to address forgetting. Some focus on preserving knowledge by maintaining consistency of certain metrics (e.g. logits or feature similarity), while others directly restricted changes to important model parameters. The architecture-based models~\cite{wang2022beef,wang2022foster,zhou2022model,hu2023dense,xu2024distribution} dynamically expanded network structures to adapt to the evolving data stream. These methods typically froze previously learned parameters and initialize new parameters to learn the knowledge of new classes. 

\subsection{Pre-Trained Model Based CIL} Recently pre-trained model based CIL has attracted rising attention due to the promising results \cite{yu2024boosting}. 
Most Pre-Trained Model-based CIL methods utilize the Parameter-Efficient Fine-Tuning (PEFT) mechanism to adapt the model efficiently while keeping the Pre-Trained model frozen. For example, SSIAT \cite{tan2024semantically} and EASE \cite{zhou2024expandable} adapted the model using adapter \cite{chen2022adaptformer} and approximate the feature of previous classes to mitigate forgetting. 
Additionally, a substantial number of methods leverage prompts. L2P \cite{wang2022learning} designed a prompt pool for incremental learning, selecting instance-specific prompts during training and inference. Most methods \cite{wang2022dualprompt,wang2023isolation,qiao2023prompt,wang2024hierarchical,liu2024compositional} developed task-specific prompts, keeping prompts for other tasks frozen during training. Instead of directly optimizing prompt parameters, DAP \cite{jung2023generating} designed prompt generators to generate more instance-specific information in prompts. However, these methods require knowledge of the task information for inference samples, necessitating task ID inference first to select the relevant prompt. And errors in task prediction can lead to inconsistencies between training and inference, reducing their performance. A few methods \cite{smith2023coda,roy2024convolutional} proposed a weighted sum of prompts that can mitigate this problem. However, they did not thoroughly explore the mechanism of weighted prompt, which limits the effectiveness of their weighted prompt strategies. In this paper, we propose a prompt aggregation strategy with concave and linear constraints which guarantee the prediction error of aggregated prompt is lower than task-specific prompt. A cyclic prompt weight strategy is also proposed to fully utilize the learned prompts to adjust the prompt weight, which further improves performance.
\begin{figure*}[t]
    \centering
	\includegraphics[width=1\linewidth]{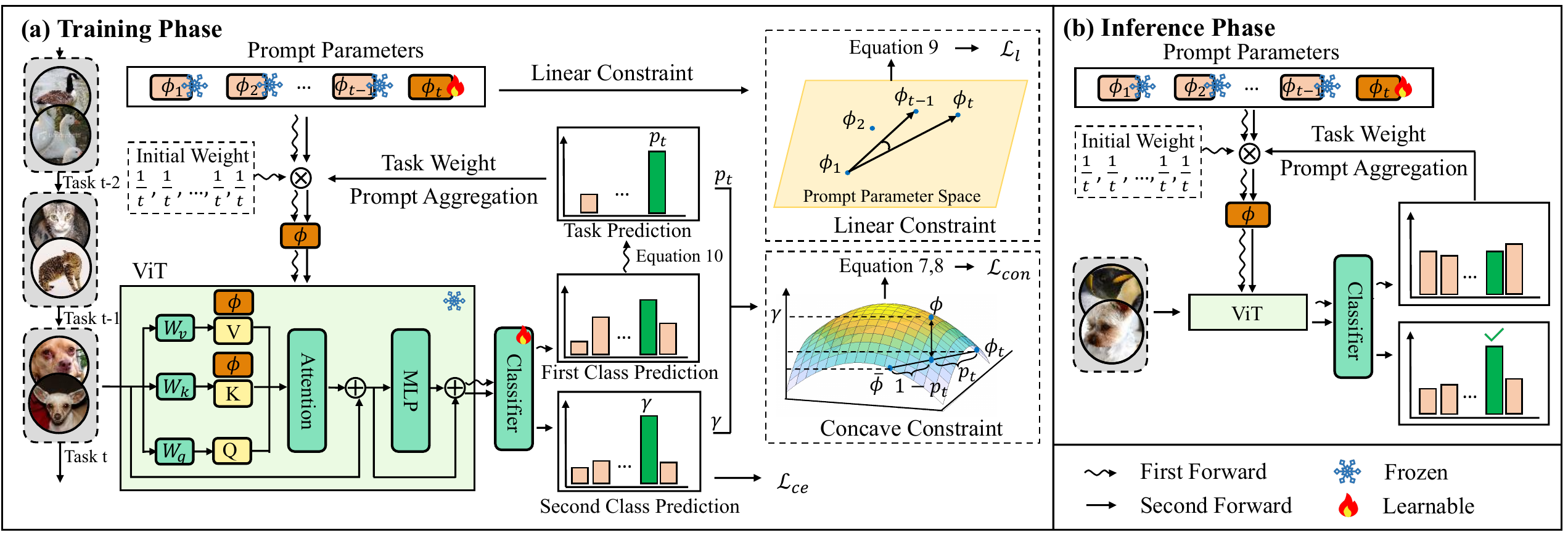}
        \caption{\label{fig:framework} The overall pipeline of our proposed CAPrompt. To overcome the inconsistency of prompts between the training phase (a) and inference phase (b), a prompt aggregation strategy is proposed. The concave and linear constraints are proposed to guarantee the prediction error of the aggregated prompt is the lower bound of using one task-specific prompt. Then, to make full use of prompts in predicting weight for prompt aggregation, we propose a cyclic prompt weight that we initiate the prompt with equal weight and cyclically predict the prompt weight. This strategy can be conducted cyclically many times to further improve performance.}

\end{figure*}

\section{Cyclic Prompt Aggregation}
\subsection{Problem Formulation}
Class Incremental Learning involves learning a model from a data stream. For a data stream with $T$ tasks, $\mathcal{D}=\{D_t\}_{t=1}^{T}$, each dataset $D_t=\{X_t, Y_t\}$ consists of input data set $X_t=\{x_{t,j}\}_{j=1}^{n_t}$ and a label set $Y_t=\{y_{t,j}\in \mathcal{C}_t\}_{j=1}^{n_t}$, where $n_t$ is the number of data in task $t$, $x_{t,j}$ represents the $j$-th image and $y_{t,j}$ represents the label. $\mathcal{C}_t$ is the label set and labels of different tasks are disjoint, that is $\mathcal{C}_i \cap \mathcal{C}_j = \emptyset (i \neq j).$ In task $t$, the model learns a mapping function $f_t: \mathbb{R}^{h \times w \times 3} \rightarrow \mathbb{R}^{l_t}$ for all seen classes, where $l_t=\sum_{j=1}^t | \mathcal{C}_j|$ is the number of classes that have been learned up to task $t$. In the pre-trained model based CIL, $f_t$ typically consists of a frozen Pre-Trained model $\theta$ (e.g. Vision Transformer), the classification head $W$, and learnable parameters $\phi_t$ (e.g. prompts). Thus, for an input image $x$, the prediction with prompt $\phi_t$ is $f_t(x,W)$, which can also be formed as $W^\top f(x,\phi_t)$. The prediction of solely using the pre-trained model is $W^\top f(x)$.

\subsection{Prompt Aggregation}

As mentioned above, existing methods need to decide which task-specific prompt to use during inference, and errors in task prediction can lead to inconsistencies between training and inference, resulting in knowledge forgetting. To mitigate this problem, we propose the Prompt Aggregation strategy which aggregates the knowledge of prompts for different tasks instead of selecting a single task-specific prompt.

Firstly, during the training of task $t$, given an image $x$ with label $y$, instead of only utilizing the task-specific prompt $\phi_t$,  we calculate the task similarity for task $i$ using the query function $f(x)$, 
\begin{equation}
    p_i = \sum_{m \in \mathcal{C}_i} {\rm Softmax}(W^\top f(x))[m],
    \label{eq:task similarity}
\end{equation}
 where $W$ consists of the maintained key feature for each seen class, ${\rm Softmax}$ represents softmax operation. Then the aggregated prompt can be calculated by:
\begin{equation}
    \phi = \sum_{i=1}^{t-1} p_i \cdot {\rm stop}(\phi_i) + p_t \cdot \phi_t.
\label{eq: weighted prompt}
\end{equation}
Here, we use stop gradient operation, ${\rm stop}$, for the prompt of previous tasks to prevent the interference of previous prompt parameters. Then the optimization function with cross-entropy loss is:  
\begin{equation}
    \mathcal{L}_{ce} = CE(W^\top f(x,\phi),y).
\end{equation}

During inference, in the first stage, we obtain the aggregated prompt according to Eq.\ref{eq:task similarity} and Eq.\ref{eq: weighted prompt}. Then in the second stage, the prediction of $x$ is given by ${\rm argmax}_m (W^\top f(x,\phi)[m])$. This prompt aggregation strategy not only ensures the consistency of prompts between training and inference but also satisfies the following theorem. 

To simplify the notation, we denote the prediction of label $y$ using prompt $\phi$, ${\rm Softmax}(W^\top f(x,\phi))[y]$ as $g(x,y,\phi)$, where $W$ represents the classification head for all classes.

\textbf{Theorem 1:} \textit{In pre-trained based CIL, for dataset $D=\{X, Y\}$, the expected error ($E_1$) of using aggregated prompt, with the weight of each prompt being the sum of probability of classes of each task, is lower than the expected error ($E_2$) of using task-specific prompts, if the prediction of each class is concave to the combination of prompts for different tasks.}

\textbf{Proof:} The expected error of the aggregated prompt is:
\begin{equation}
    E_1 =\mathbb{E}_x[{\rm -log} \ g(x,y,\sum_{i=1}^t p_i \cdot \phi_i)].
\end{equation}
The expected error using task-specific prompts is:
\begin{equation}
    E_2 = \mathbb{E}_x[{\rm -log} \sum_{i=1}^t p_i  \cdot g(x,y, \phi_i)],
\end{equation}
where $p_i$ is the task similarity in Eq.\ref{eq:task similarity}, represents the probability of task $i$ being the most similar task. When the prediction of the label $y$ is concave in the prompt parameter space, according to Jensen Inequality:
\begin{equation}
    g(x,y,\sum_{i=1}^t p_i \cdot \phi_i) \ge \sum_{i=1}^t p_i  \cdot g(x,y, \phi_i),
\end{equation}
then $E_1 \le E_2$. Finish the proof (detailed in Supplementary).

This theorem demonstrates that the aggregated prompt yields a lower prediction error than using one task-specific prompt under the concave condition. Thus, to satisfy this condition, we propose the following \textbf{Concave Constraint} and \textbf{Linear Constraint}.

\textbf{Concave Constraint.} An intuitive way to meet with the concave condition is to directly constrain the prediction with different prompts during training. Thus for an image $x$ with label $h$, the Concave Constraint for task $t$ is defined as:
\begin{equation}
\begin{split}
    \delta =p_t \cdot g(x,y, \phi_t) 
    &+ (1-p_t) \cdot g(x,y,\sum_{i=1}^{t-1} \frac{p_i  }{1-p_t}\phi_i) \\ &- g(x,y,\sum_{i=1}^t p_i \cdot \phi_i),
\end{split}
\label{eq:delta}
\end{equation}
\begin{equation}
    \mathcal{L}_{con} = {\rm max} (\delta,0).
\end{equation}

By employing the concave loss in each task, we explicitly constrain the prompt to satisfy the concave condition.

\begin{figure}[h]
    \centering
	\includegraphics[width=0.95\linewidth]{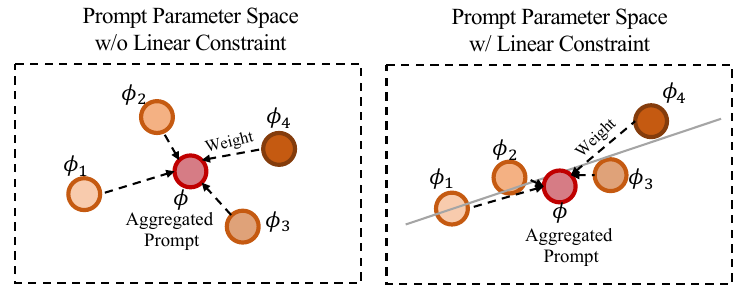}
        \caption{\label{fig:motivation2} Motivation of Linear Constraint.}
\end{figure}
\textbf{Linear Constraint.}
Though we propose a concave constraint to ensure the model satisfies the concave condition in Theorem 1, this condition cannot be exactly met due to the large number of parameters. Moreover, the concave condition in Theorem.\ref{alg:algorithm} focuses on the combination of learned prompts for different tasks rather than the whole prompt parameters space. Thus we tend to restrict the learning of prompts to make the concave condition easier to meet. As illustrated in Fig.\ref{fig:motivation2}, due to the aggregated prompt being the linear combination of different prompts, we propose a linear constraint to ensure that the linear combination of prompts is also close to the learned prompts. Specifically, the linear constraint forces the prompts of different tasks to lie in a linear direction by maximizing the cosine similarity of the prompt direction relative to the prompt of the first task. The linear constraint is formulated as:
\begin{equation}
    \mathcal{L}_l = 1-{\rm sim}(\phi_t-\phi_1,\phi_{t-1}-\phi_1),
    \label{eq:linear constraint}
\end{equation}
where ${\rm sim}$ represents the cosine similarity.

\begin{algorithm}[tb]
\caption{Cyclic Prompt Aggregation algorithm}
\label{alg:algorithm}
\textbf{Training:}\\
\textbf{Input}: Data stream $\mathcal{D}=\{D_t\}_{t=1}^{T}$, a pre-trained ViT $f(\cdot)$.\\
\textbf{Output}: Prompts $\phi_1,...,\phi_T$ and classification head $W$.
\begin{algorithmic}[1] 
\FOR{$t$ in 1 : $T$} 
\STATE Set initial weight $p_1^1 = ... = p_t^1 = \frac{1}{t}$.
\STATE Get aggregated prompt $\phi^1$ by Eq.\ref{eq: weighted prompt} with $p_1^1,...,p_t^1$.
\STATE Get task probability $p_1^2,..,p_t^2$ by Eq.\ref{eq:task similarity2} with $\phi^1$.
\STATE Get aggregated prompt $\phi^2$ by Eq.\ref{eq: weighted prompt} with $p_1^2,...,p_t^2$.
\STATE Compute the loss $\mathcal{L}_{ce}+\mathcal{L}_{con}+\mathcal{L}_{l}$ with $\phi^2$ and update parameters.
\ENDFOR
\end{algorithmic}

\textbf{Inference after training on task $t$:}\\
\textbf{Input}: Input image $x$, a pre-trained ViT $f(\cdot)$, prompts $\phi_1,...,\phi_t$, classification head $W$, number of cycles $num$.\\
\textbf{Output}: Class number $m$.
\begin{algorithmic}[1] 
\STATE Let $n = 1$.
\STATE Set initial weight $p_1^n = ... = p_t^n = \frac{1}{t}$.
\FOR{$n$ in 1 : $num$} 
\STATE Get aggregated prompt $\phi^{n}$ by Eq.\ref{eq: weighted prompt} with $p_1^{n},..,p_t^{n}$.
\STATE Get task probability $p_1^{n+1},..,p_t^{n+1}$ by Eq.\ref{eq:task similarity2} with $\phi^n$.
\ENDFOR
\STATE $m = {\rm argmax}_m (W^\top f(x,\phi^{num})[m])$.
\end{algorithmic}
\end{algorithm}

\begin{table*}[t]
  \centering
  {
  \setlength{\tabcolsep}{5.1pt}
  \renewcommand{\arraystretch}{1.25}
    {
    
    \begin{tabular}{l|c|c|c|c|c|c|c}
    \hline
    \multirow{2}{*}{Methods}  & \multirow{2}{*}{Publication} & \multicolumn{2}{c|}{CIFAR-100} & \multicolumn{2}{c|}{ImageNet-R} & \multicolumn{2}{c}{CUB200} \\

    \cline{3-8} &  & $ACC$  & $AF$  & $ACC$   & $AF$   & $ACC$  & $AF$    \\
    \hline
    L2P & CVPR'22 & 83.06 $\pm$ 0.40 & 5.95 $\pm$ 0.47 & 63.22 $\pm$ 0.34  & 7.05 $\pm$ 0.37 & 70.88 $\pm$ 0.03 & 6.04 $\pm$ 0.39 \\
    DualPrompt & ECCV'22 & 85.11 $\pm$ 0.29 & 5.74 $\pm$ 0.46 & 69.14 $\pm$ 0.16 &4.55 $\pm$ 0.01  & 76.54 $\pm$ 0.15  & 5.67 $\pm$ 0.20 \\
    CODA-Prompt & CVPR'23 & 86.76 $\pm $ 0.22 &6.36 $\pm$ 0.28 & 73.71 $\pm$ 0.25 &5.11 $\pm$ 0.65  &73.71 $\pm$ 0.92 & 7.49 $\pm$ 0.20 \\
    
    HiDe-Prompt & NeurIPS'23 &  \textcolor{blue}{92.64 $\pm$ 0.22}& \textcolor{blue}{1.92 $\pm$ 0.23} & 75.66 $\pm$ 0.21&  \textcolor{red}{\textbf{2.88 $\pm$ 0.19}}& \textcolor{blue}{86.84 $\pm$ 0.14}& \textcolor{blue}{2.07 $\pm$ 0.05} \\
    EvoPrompt & AAAI'24  & 87.57 $\pm$ 0.40 & 5.50 $\pm$ 0.44 & 76.49 $\pm$ 0.27& 3.57 $\pm$ 0.13&79.88 $\pm$ 0.31 & 9.40 $\pm$ 0.81\\
    ConvPrompt & CVPR'24 & 88.86 $\pm$ 0.21 & 3.37 $\pm$ 0.24 & \textcolor{blue}{77.94 $\pm$ 0.06} & 3.43 $\pm$ 0.16 &81.08 $\pm$ 0.52  &5.97 $\pm$ 0.67 \\
    CPrompt & CVPR'24 & 87.83 $\pm$ 0.13 & 4.88 $\pm$ 0.02& 76.70 $\pm$ 0.23& 6.08 $\pm$ 0.19 & 82.69 $\pm$ 0.43& 5.30 $\pm$ 0.35\\
    \hline
    \rowcolor{gray!25}Ours & This Paper & \textcolor{red}{\textbf{95.52} $\pm$ \textbf{0.12}}& \textcolor{red}{\textbf{1.76} $\pm$ \textbf{0.20}}&\textcolor{red}{\textbf{79.93} $\pm$ \textbf{0.19}}& \textcolor{blue}{3.37 $\pm$ 0.41} & \textcolor{red}{\textbf{88.99} $\pm$ \textbf{0.15}} & \textcolor{red}{\textbf{1.46} $\pm$ \textbf{0.19}}\\
    \hline
    \end{tabular}
    }
    \caption{\label{tab:main results}Comparison of different continual learning methods on various dataset settings with ImageNet pre-trained ViT. We report results averaged over 3 trials. The best results are marked in red. The second best results are marked in blue.}
    }
    
\end{table*}

In conclusion, by employing prompt aggregation in both training and inference, we eliminate the task prediction and ensure the consistency of prompts, thus mitigating knowledge forgetting. Additionally, the concave and linear constraints are designed to ensure that the aggregated prompt yields a lower prediction error than using a single task-specific prompt.
\subsection{Cyclic Prompt Weight}
Though the proposed Prompt Aggregation enhances the consistency of prompts, the aggregation weight of prompts for each task is obtained by the pre-trained model without prompts which neglects the knowledge of prompts. To fully utilize the learned prompts, we propose a cyclic prompt weight prediction strategy. In the first stage during training and inference, we get the aggregated prompts $\phi^1$ by Eq.\ref{eq: weighted prompt} with equal weight for each prompt, $p_1^1 = p_2^1 = ... = p_t^1 = \frac{1}{t}$, and get the class probability prediction. Then the Eq.\ref{eq:task similarity} can be formed as follows to calculate the new task probability for cycle $n+1$ with aggregated prompt $n$:
\begin{equation}
    p_i^{(n+1)} = \sum_{m \in \mathcal{C}_i} {\rm Softmax}(W^\top f(x,\phi^n))[m].
    \label{eq:task similarity2}
\end{equation}
This task probability serves as the weight to calculate the new aggregated prompts $\phi^2$ by Eq.\ref{eq: weighted prompt}. This updated prompt $\phi^2$ is then used to prompt the model in the second stage. 

Moreover, apart from the two-stage inference paradigm, our method can be conducted cyclically multiple times to obtain more accurate aggregation weights, further improving performance.

\subsection{Overall Optimization}
The overall pipeline of our proposed CAPrompt is shown in Fig.\ref{fig:framework} and the training and inference process is detailed in Algorithm \ref{alg:algorithm}. During training, we perform in a two-stage manner as existing methods. Initially, we assign equal task probabilities to different prompts, then get the aggregated prompts $\phi^1$ and the new task probability using Eq.\ref{eq: weighted prompt} and Eq.\ref{eq:task similarity2}. The updated aggregated prompt $\phi^2$ is then calculated through Eq.\ref{eq: weighted prompt}. The loss function with hyperparameters $\alpha$ and $\beta$ is calculated as:
\begin{equation}
    \mathcal{L}=\mathcal{L}_{ce} + \alpha \cdot \mathcal{L}_{con} + \beta \cdot \mathcal{L}_{l}.
\end{equation}

During inference, as shown in Algorithm \ref{alg:algorithm}, we also initialize equal task probabilities for different prompts. The aggregated prompt and task probabilities are then calculated cyclically over a specified number of cycles $num$. The final aggregated prompt $\phi^{num}$ is used to prompt the model to get the prediction. Note that existing methods rely on the key-query matching strategy to select the relevant prompt and require two forward passes through the network. Thus, the computation cost for CAPrompt is similar to existing methods when $num=2$.

\begin{figure*}[t]
    \centering
	\includegraphics[width=0.93\linewidth]{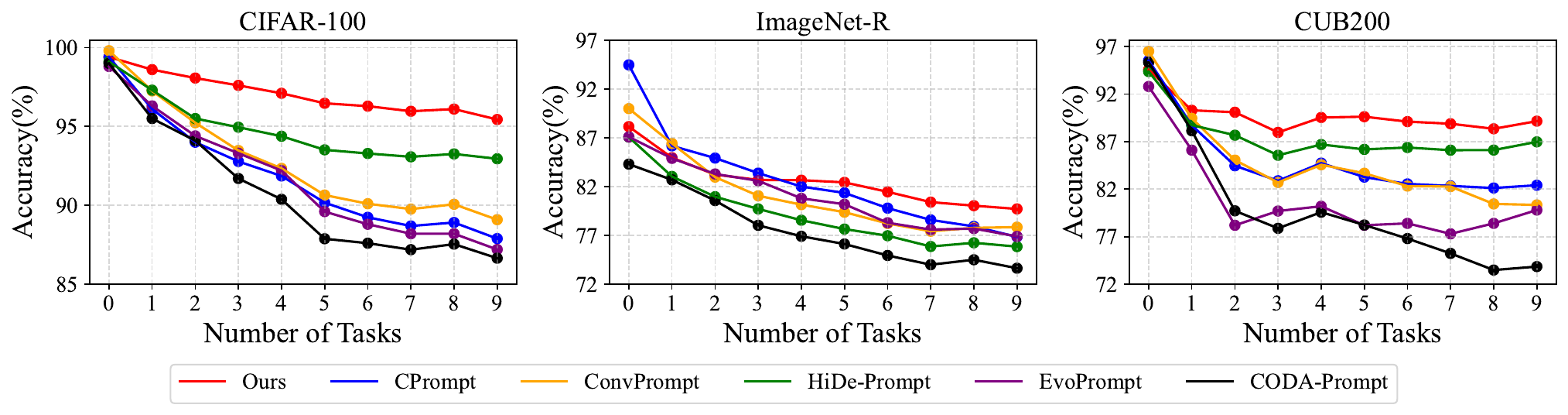}
        \caption{The complete classification accuracy of different methods on each task.\label{fig:acc_curve} }

\end{figure*}

\section{Experiments}
\subsection{Experimental Details}
\subsubsection{Datasets.} We follow existing works \cite{wang2024hierarchical} to evaluate our proposed method on three public datasets, CIFAR-100~\cite{krizhevsky2009learning}, ImageNet-R~\cite{hendrycks2021many}, and CUB200~\cite{wah2011caltech}. CIFAR-100, ImageNet-R and CUB200 comprise 100, 200 and 200 classes respectively. These datasets are splited into 10 tasks with disjoint classes for incremental learning.

\subsubsection{Evaluation Metrics.}
Following previous works~\cite{wang2024hierarchical}, we use final accuracy ($ACC$) and average forgetting ($AF$) for evaluation. $ACC$ represents the average accuracy of all the classes that have already been learned. $AF$ estimates the forgetting of previous tasks by calculating the average performance degradation of each classes in different tasks during incremental learning.

\subsubsection{Comparison Methods.}
We compare our methods with various prompt-based pre-trained model based CIL methods, such as L2P \cite{wang2022learning}
, DualPrompt \cite{wang2022dualprompt}, CODA-Prompt \cite{smith2023coda}, HiDe-Prompt \cite{wang2024hierarchical}, EvoPrompt \cite{kurniawan2024evolving}, ConvPrompt \cite{gao2024consistent} and Cprompt \cite{roy2024convolutional}. We also include existing methods using other PEFT, such as InfLoRA \cite{liang2024inflora}, EASE \cite{zhou2024expandable} and SSIAT \cite{tan2024semantically}.

\subsubsection{Implementation Details.}
For all experiments, we use the ImageNet21K pre-trained ViT-B/16 as the backbone. The parameters are optimized by an Adam optimizer with an initial learning rate of 3e-3 and a batch size of 24. Prefix Tuning Prompts \cite{wang2022dualprompt}, with the prompt length $L_p=10$ are inserted into all layers for CIFAR-100, CUB200, and into the first nine layers for ImageNet-R with $L_p=20$. The prompt parameters for each task are initialized using the prompts from the previous task. The weighting parameters of different losses are $\alpha=5$ and $\beta=0.2$. The cyclic number $num$ is set to 2 without special explanations for a similar computation cost compared to existing methods. Following HiDe-Prompt, we also maintain one feature per class and replay this feature in subsequent incremental learning tasks to mitigate the forgetting of the classification head. For comparison methods, we reimplement their results according to their released code and the experimental setting reported in their paper. All results are the average performance across 3 runs. 

\subsection{Comparison with SOTA}
\subsubsection{Main Results.}
Table.\ref{tab:main results} represents the results of final accuracy on various datasets. Compared to other prompt-based methods, our CAPrompt achieves the highest accuracy. Specifically, our method achieves final accuracies of \textbf{95.52\%}, \textbf{79.93\%}, and \textbf{88.99\%} on CIFAR-100, ImageNet-R, and CUB200 datasets with the performance gain of 2.88\%, 1.99\%, and 2.15\% compared to the second best player. In terms of average forgetting, our results demonstrate the lowest forgetting rate on CIFAR-100 and CUB200. Although HiDe-Prompt exhibits slightly lower forgetting on ImageNet-R, its knowledge acquisition is limited, leading to inferior final accuracy results. Furthermore, we compare our method with the latest LoRA-based and Adapter-based methods in Table.\ref{tab:other PEFT}. Results show that our method also achieves the highest final accuracy and the lowest average forgetting demonstrating its effectiveness.

\begin{table}[t]
  \centering
  {
  \renewcommand{\arraystretch}{1.3}

    \begin{tabular}{l|c|c|c|c|c|c}
    \hline
    \multirow{2}{*}{Methods}  &\multicolumn{2}{c|}{CIFAR-100} & \multicolumn{2}{c|}{ImageNet-R} & \multicolumn{2}{c}{CUB200} \\

    \cline{2-7}   & $ACC$  & $AF$  & $ACC$   & $AF$   & $ACC$  & $AF$   \\
    \hline

    InfLoRA & 86.60 &4.87& 74.77 & 6.60 & 77.45 & 4.93 \\
    EASE &  87.58 & 6.55 & 76.80& \textcolor{blue}{5.07}& 86.74 & \textcolor{blue}{2.63} \\
    SSIAT &  \textcolor{blue}{90.71} & \textcolor{blue}{4.89} &\textcolor{blue}{79.37} & {6.96} & \textcolor{blue}{87.06} & 7.63\\
    \hline
    \rowcolor{gray!25}Ours   & \textcolor{red}{\textbf{95.52}}& \textcolor{red}{\textbf{1.76}} & \textcolor{red}{\textbf{79.93}}& \textcolor{red}{\textbf{3.37}} & \textcolor{red}{\textbf{88.99}}& \textcolor{red}{\textbf{1.46}}   \\
    \hline
    \end{tabular}

    \caption{\label{tab:other PEFT} Comparison of LoRA-based and Adapter-based methods on various dataset settings.}
    }
    
\end{table}

The outstanding results of our method are attributed to the prompt aggregation strategy, which ensures the consistency of prompts between training and inference. While CODA-Prompt proposed a weighted sum of prompts with an attention-based component-weighting scheme for different prompts, it lacks constraints on the learning of prompt parameters. In contrast, our CAPrompt employs concave and linear constraints, ensuring that the aggregated prompt yields lower prediction errors than using a single task-specific prompt. Additionally, the proposed cyclic prompt weight strategy fully utilizes the knowledge learned by prompts, further improving performance.

\subsubsection{Accuracy Curve.}
To present our results in detail, we present the final accuracy of different methods after different tasks in Fig.\ref{fig:acc_curve}. Notably, with similar accuracy for the initial task, our method achieves the best results across subsequent tasks. This performance gain can be attributed to the robust prompt aggregation strategy and the cyclic prompt weight strategy which effectively leverages the information captured by the prompts.
\begin{table}[t]
  \centering
  {
  \renewcommand{\arraystretch}{1.3}

    \begin{tabular}{l|c|c}
    \hline
    \multirow{2}{*}{Methods}  & \multicolumn{2}{c}{ImageNet-R} \\

    \cline{2-3}&    $ACC$   & $AF$   \\
    \hline
     Base  &    74.51  &  3.01  \\
    + Aggregation  &   78.28 & 4.41 \\
    + Aggregation + Cyclic &  79.16 & 4.52 \\
    + Aggregation + Cyclic + $\mathcal{L}_{con}$&   79.77  & 3.70 \\
    + Aggregation + Cyclic + $\mathcal{L}_{con}$ + $\mathcal{L}_{l}$&  79.93  & 3.37 \\
    
    \hline
    \end{tabular}

    \caption{\label{tab:ablation} Ablation study of different components.}
    }
    
\end{table}

\begin{figure*}[t]
    \centering
	\includegraphics[width=1\linewidth]{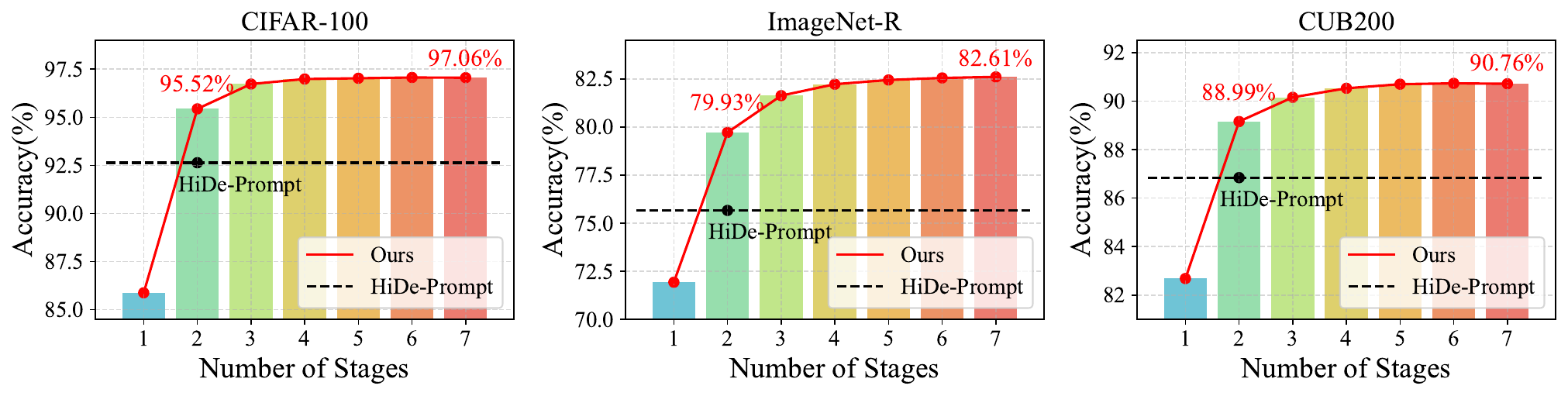}
        \caption{\label{fig:multi} Accuracy of the proposed method increases with the number of cycles ($num$).}
\end{figure*}

\subsection{Ablation Study}
\subsubsection{Effectiveness of Different Components.} The ablation results on ImageNet-R are presented in Table.\ref{tab:ablation}. The base model uses task-specific Prefix Tuning Prompts for the first nine layers, with prompt parameters initialized from previous tasks. Our method comprises two strategies and two losses: prompt aggregation, cyclic prompt weight, $\mathcal{L}_{con}$, and $\mathcal{L}_{l}$. Table.\ref{tab:ablation} demonstrates that incorporating prompt aggregation leads to a significant improvement in final accuracy. This is because the prompt aggregation ensures consistency of prompts between the training and inference, thus mitigating knowledge forgetting. Moreover, the aggregated prompt combines similar knowledge from multiple classes and enhances knowledge acquisition. Additionally, the cyclic prompt weight strategy further improves the accuracy by 0.88\%, due to its effective utilization of prompts for weight determination. These improvements in knowledge acquisition increase the difficulty of anti-forgetting, thus leading to a slight increase in the metric of average forgetting. Moreover, the concave and linear constraints each contribute to a 0.61\% and 0.16\% increase in final accuracy and a 0.82\% and 0.33\% reduction in average forgetting, respectively. This highlights their effectiveness in constraining prompt learning and mitigating knowledge forgetting.

\begin{figure}[t]
    \centering
	\includegraphics[width=1\linewidth]{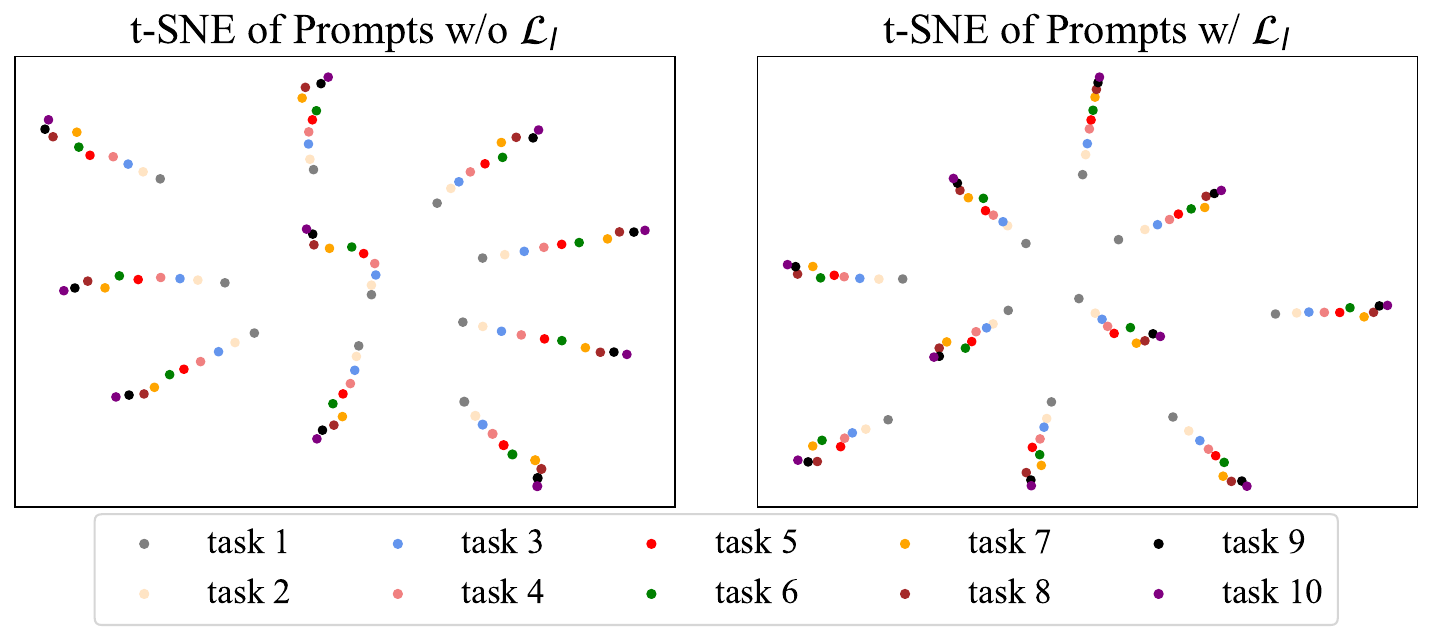}
        \caption{\label{fig:linear} t-SNE visualization of prompts w/o $\mathcal{L}_{l}$ and w/ $\mathcal{L}_{l}$.}
\end{figure}

\subsubsection{Linear Constraint.} To evaluate the impact of $\mathcal{L}_{l}$, we visualize the prompts of layer one for key tokens after training on the ImageNet-R dataset in Fig.\ref{fig:linear}. The number of prompts is $L_p / 2=10$ for key tokens. The visualization results show that applying $\mathcal{L}_{l}$ effectively aligns the prompts of different tasks in a linear direction and brings them closer together. This alignment facilitates prompt aggregation by constraining the possible locations of aggregated prompts, which simplifies the training process to meet the concave condition described in Theorem 1. Consequently, this alignment ensures that the aggregated prompt achieves a lower prediction error compared to using a single task-specific prompt.

\subsubsection{Cyclic Prompt Weight.} 
As described in the method section, our approach allows for multiple cycles to refine the aggregation weight. For a fair comparison, we set the cyclic number $num=2$ for the results shown in Table.\ref{tab:main results}, Table.\ref{tab:other PEFT} and Table.\ref{tab:ablation}. In this section, we evaluate the impact of performing additional cycles on prediction accuracy and present the results in Fig.\ref{fig:multi}. Our findings reveal that, with two cycles ($num=2$), our method achieves superior accuracy compared to the second-best method, HiDe-Prompt. Furthermore, by iterating the cyclic prompt weight adjustment over multiple stages, we observe an additional performance improvement of 1.5\%-2\%. This enhancement is attributed to the iterative refinement of aggregation weights, which leads to progressively more accurate prompt combinations and consequently better overall performance.

\begin{figure}[t]
    \centering
	\includegraphics[width=0.83\linewidth]{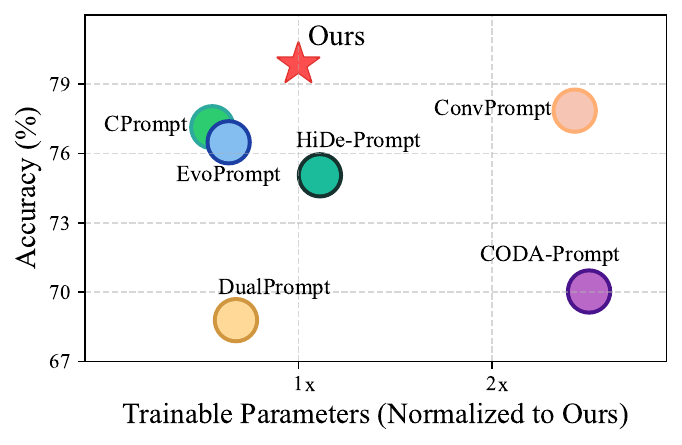}
        \caption{\label{fig:acc-param} The accuracy and number of trainable parameters for different methods on the ImageNet-R dataset. All trainable parameters are normalized to our method.}
\end{figure}

\subsubsection{Parameters Comparison.}
Fig.\ref{fig:acc-param} presents the accuracy and the number of trainable parameters for different methods on ImageNet-R datasets, with all trainable parameters normalized according to our method. CODA-Prompt proposed an attention-based component-weighting scheme, and ConvPrompt employs a convolutional prompt generator, both requiring more parameters. Although CPrompt utilizes the fewest parameters, it sequentially prepends prompts to the input tokens, increasing the number of tokens and the computational cost. Our method requires a similar number of trainable parameters compared to other methods but achieves the highest accuracy performance. This superior performance is attributed to our prompt aggregation and cyclic prompt weight strategies, which make full use of the knowledge embedded in the prompts.

\section{Conclusion}

In this paper, we present the Cyclic Prompt Aggregation (CAPrompt) method to address the challenges of prompt inconsistency in task ID prediction in Class Incremental Learning (CIL). By aggregating the knowledge of different prompts, CAPrompt eliminates the need for task ID prediction, thereby reducing inconsistencies between training and inference prompts and mitigating knowledge forgetting. The concave constraint and linear constraint are proposed to ensure the aggregated prompt yields lower error than using one task-specific prompt. Additionally, the cyclic weight prediction strategy is proposed to cyclically adjust the prompt weights to more accurate ones, further enhancing performance. Extensive experiments on various datasets demonstrated that CAPrompt significantly outperforms the state-of-the-art methods. Our results underscore the effectiveness of prompt aggregation and the importance of cyclically utilizing the learned prompts in CIL. 

\bigskip

\noindent \textbf{Acknowledgment.} This work was supported by the National Natural Science Foundation of China (62376011).
\bibliography{aaai25}

\clearpage

\section{Influence of Hyperparameters}
\begin{figure}[h]
    \begin{center}
	\includegraphics[width=1\linewidth]{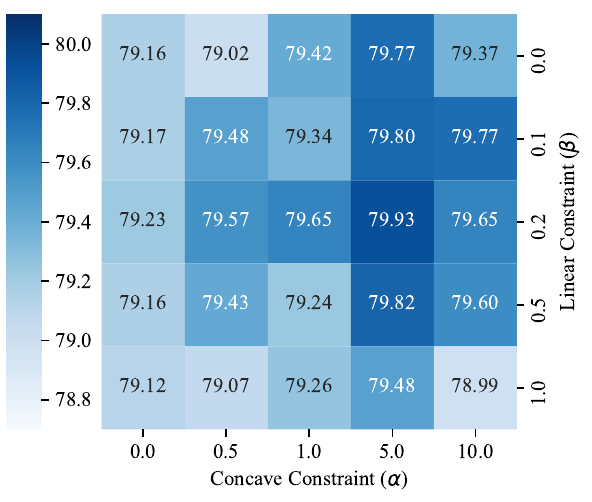}
        \caption{\label{fig:ablation}Influence of different hyperparameters. }
    \end{center}

\end{figure}

In this section, we explore the impact of hyperparameters on the ImageNet-R dataset. Our method includes two hyperparameters, $\alpha$ and $\beta$, which represent the weights of $\mathcal{L}_{con}$ and $\mathcal{L}_l$, respectively. The results for different values of $\alpha$ and $\beta$ are shown in Fig.\ref{fig:ablation}. $\mathcal{L}_{con}$ and $\mathcal{L}_l$ are loss functions designed to constrain prompt learning to satisfy the concave condition. While higher weights for these loss functions can better ensure that the concave condition is met, they also limit the learning capability of the prompts. Therefore, we empirically chose $\alpha=5$ and $\beta=0.2$ for the best performance.

\section{Learning Capability Influence of $\mathcal{L}_l$}

As illustrated in the paper, we tend to restrict the learning of prompts to make it easier to satisfy the concave condition. To this end, we propose a linear constraint, $\mathcal{L}_l$, that forces the prompts of different tasks to align in a linear direction. This constraint potentially limit the learning capability of prompts. We explore its influence by reporting results where test images are forwarded using the prompt associated with the ground truth task ID, reflecting the model's learning capability. Results in Table.\ref{tab:inc} show that the performance of with and without $\mathcal{L}_l$ is similar, indicating that $\mathcal{L}_l$ does not explicitly limit the learning capability of prompts.

\section{Different Pre-Trained Models.} In this section, we compare different prompt-based methods \cite{wang2022learning,wang2022dualprompt,smith2023coda,kurniawan2024evolving,gao2024consistent,wang2024hierarchical} using different self-supervised pre-trained models on CIFAR-100, ImageNet-R, and CUB200 in Table.\ref{tab:pretrained backbone}. We reimplement all comparison methods with their code. Due to the reimplemented results of HiDe-Prompt on CIFAR-100 and ImageNet are lower than those reported in the paper, we directly use their reported results. It can be observed that, although the performance with self-supervised pre-trained models is lower than with supervised ones, our CAPrompt method consistently outperforms existing approaches. This demonstrates the effectiveness of CAPrompt in achieving a better balance between knowledge acquisition and forgetting, even when leveraging self-supervised pre-trained models.

\begin{table}[t]
  \begin{center}
  {
  \renewcommand{\arraystretch}{1.3}

    \begin{tabular}{l|c|c|c}
    \hline
    Methods  &CIFAR-100& ImageNet-R & CUB200 \\
    \hline

    Ours w/o $\mathcal{L}_l$ & 97.85 $\pm$ 0.02 &84.76 $\pm$ 0.12& 91.07 $\pm$ 0.25 \\
    Ours  & 97.90 $\pm$ 0.01 &84.78 $\pm$ 0.18& 90.99 $\pm$ 0.15 \\
    \hline
    
    \end{tabular}

    \caption{\label{tab:inc} Performance of different methods with ground truth task ID.}
    }
    \end{center}
    
\end{table}

\begin{table*}[ht]
  \begin{center}
  {
  \renewcommand{\arraystretch}{1.3}

    \begin{tabular}{c|l|c|c|c|c|c|c}
    \hline
   \multirow{2}{*}{} & \multirow{2}{*}{Methods} & \multicolumn{2}{c|}{CIFAR-100} & \multicolumn{2}{c|}{ImageNet-R} & \multicolumn{2}{c}{CUB200}\\

    \cline{3-8}&  & $ACC$  & $AF$  & $ACC$   & $AF$& $ACC$   & $AF$   \\
    \hline
    \multirow{8}{*}{\rotatebox{90}{iBOT-1k}} & L2P  & 66.39 $\pm$ 0.50  & 18.71 $\pm$ 0.33 & 61.19 $\pm$ 0.65  & 4.93 $\pm$ 0.96  & 40.82 $\pm$ 0.63 & 9.29 $\pm$ 1.03\\
    & DualPrompt  & 69.53 $\pm$ 0.80 & 15.79 $\pm$ 0.46 & 61.95 $\pm$ 0.59 &  5.43 $\pm$ 0.45 & 41.81 $\pm$ 0.55 & 11.40 $\pm$ 1.44 \\
    & CODA-Prompt &  79.29 $\pm$ 0.48 &9.01 $\pm$ 0.35& 67.17 $\pm$ 0.29 & 7.00 $\pm$ 0.39& 50.87 $\pm$ 0.48& 10.00 $\pm$ 0.52\\
    & EvoPrompt &  79.55 $\pm$ 0.07 &10.51 $\pm$ 0.46& 65.08 $\pm$ 0.19 & 6.47 $\pm$ 0.45& 58.76 $\pm$ 1.50& 15.63 $\pm$ 1.80\\
    &ConvPrompt  &  79.85 $\pm$ 0.11 &8.04 $\pm$ 0.12& 71.31 $\pm$ 0.44 & 5.66 $\pm$ 0.46& 53.18 $\pm$ 1.27& 10.87 $\pm$ 2.38\\
    &CPrompt  & 81.15 $\pm$ 0.18 &8.71 $\pm$ 0.37&  70.15 $\pm$ 0.33 &8.67 $\pm$ 0.58&61.01 $\pm$ 0.33& 13.14 $\pm$ 0.26\\
    & HiDe-Prompt  & \textcolor{blue}{93.48 $\pm$ 0.11} & \textcolor{red}{\textbf{1.00 $\pm$ 0.24}}& \textcolor{blue}{71.33 $\pm$ 0.21} & \textcolor{red}{\textbf{2.79 $\pm$ 0.26}} & \textcolor{blue}{73.15 $\pm$ 0.53} & \textcolor{blue}{3.52 $\pm$ 0.45}\\
    &\cellcolor{gray!25}Ours  &\cellcolor{gray!25}\textcolor{red}{\textbf{93.81 $\pm$ 0.05}} &\cellcolor{gray!25}\textcolor{blue}{1.57 $\pm$ 0.03}&\cellcolor{gray!25}\textcolor{red}{\textbf{74.50 $\pm$ 0.17}} & \cellcolor{gray!25}\textcolor{blue}{2.94 $\pm$ 0.26} & \cellcolor{gray!25}\textcolor{red}{\textbf{80.42 $\pm$ 0.30}} & \cellcolor{gray!25}\textcolor{red}{\textbf{1.79 $\pm$ 0.21}}\\
    \hline
    \multirow{8}{*}{\rotatebox{90}{DINO-1K}} & L2P  & 63.65 $\pm$ 1.64 & 15.40 $\pm$ 1.76 &  57.55 $\pm$ 0.17 & 4.58 $\pm$ 0.49 & 39.76 $\pm$ 0.62 & 8.55 $\pm$ 0.99\\
    & DualPrompt  & 67.06 $\pm$ 0.54 & 14.80 $\pm$ 1.16 &  58.78 $\pm$ 0.16 & 5.86 $\pm$ 0.29 &42.27 $\pm$ 0.80 & 9.88 $\pm$ 0.88 \\
    & CODA-Prompt & 76.87 $\pm$ 0.22& 8.76 $\pm$ 0.36& 64.02 $\pm$ 0.07 &7.14$\pm$ 0.44& 51.09 $\pm$ 0.32 & 8.88 $\pm$ 0.46\\
    
    & EvoPrompt & 76.55 $\pm$ 0.34& 12.13 $\pm$ 0.39& 61.73 $\pm$ 0.09 &8.48$\pm$ 0.49& 58.86 $\pm$ 1.39 & 14.90 $\pm$ 1.11\\
    &ConvPrompt  & 77.29 $\pm$ 0.51& 7.90 $\pm$ 0.98& \textcolor{blue}{68.81 $\pm$ 0.15} &5.86 $\pm$ 0.40& 53.35 $\pm$ 0.81 & 12.94 $\pm$ 0.28 \\
    &CPrompt  & 77.96 $\pm$ 0.33& 9.98 $\pm$ 0.44 & 65.67 $\pm$ 0.15 & 9.44 $\pm$ 0.21& 58.70 $\pm$ 0.89& 15.07 $\pm$ 0.81\\
    & HiDe-Prompt  & \textcolor{blue}{92.51 $\pm$ 0.11}  &  \textcolor{red}{\textbf{0.99 $\pm$ 0.21}} &  68.11 $\pm$ 0.18 & \textcolor{red}{\textbf{3.11 $\pm$ 0.17}} & \textcolor{blue}{72.97 $\pm$ 0.66} & \textcolor{blue}{3.92 $\pm$ 0.39} \\
    &\cellcolor{gray!25}Ours  &\cellcolor{gray!25}\textcolor{red}{\textbf{92.70 $\pm$ 0.01}} &\cellcolor{gray!25}\textcolor{blue}{1.73 $\pm$ 0.09} &\cellcolor{gray!25}\textcolor{red}{\textbf{71.33 $\pm$ 0.21}} & \cellcolor{gray!25}\textcolor{blue}{3.76 $\pm$ 0.21}  &\cellcolor{gray!25}\textcolor{red}{\textbf{78.55 $\pm$ 0.42}} & \cellcolor{gray!25}\textcolor{red}{\textbf{2.00 $\pm$ 0.36}} \\
    \hline
    \end{tabular}

    \caption{\label{tab:pretrained backbone} Comparison of continual learning methods on iBOT-1k and DINO-1k pre-trained models. Best results are marked in \textcolor{red}{\textbf{red}}. The second best results are marked in \textcolor{blue}{blue}.
    }
    }
    \end{center}
    
\end{table*}

\section{Proof of Theorem 1}

\textbf{Theorem 1:} \textit{In pre-trained based CIL, for dataset $D=\{X, Y\}$, the expected error ($E_1$) of using aggregated prompt, with the weight of each prompt being the sum of probability of classes of each task, is lower than the expected error ($E_2$) of using task-specific prompts, if the prediction of each class is concave to the combination of prompts for different tasks.}

\textbf{Proof:}
Given the pre-trained model $\theta$, prompt parameters of $T$ different tasks, $\Phi=\{\phi_1,...,\phi_T\}$. As illustrated in \cite{wang2024hierarchical}, for an input $x$ of task $ h$ with label $y$, a general goal of the CIL is to learn $P(x \in X_{h,y}|\theta, \Phi)$ where $X_{h,y}$ represents the domain of class $y$ in task $h$. Following \cite{wang2024hierarchical}, we also discuss the prediction error in the context of widely-used cross-entropy loss. Specifically, the expected error ($E$) is formed as:
\begin{equation}
    E = \mathbb{E}_x[{\rm -log} \ P(x \in X_{h,y}|\theta, \Phi)]
\end{equation}
For the existing methods that use task-specific prompts, they perform in a two-stage manner. Firstly, they predict the task ID of the image, denoted as $P(x \ \hat{\in} \ X_{i}|\theta, \Phi)$. Then they get the final prediction with the relevant prompt, denoted as $P(x \in X_{h,y}|x \ \hat{\in} \ X_{i}, \theta, \Phi)$. Thus, their expected error ($E_2$) can be formed as follows based on Bayes' Rule:
 
\begin{equation}
\begin{aligned}
   E_2= &\mathbb{E}_x[{\rm -log}P(x \in X_{h,y}|\theta, \Phi) ]
   \\ =   &\mathbb{E}_x[{\rm -log}\sum_{i=1}^{T} P(x \ \hat{\in} \ X_{i}|\theta, \Phi) P(x \in X_{h,y}|x \ \hat{\in} \ X_{i}, \theta, \Phi) ]
   \\ = & \mathbb{E}_x[{\rm -log} \sum_{i=1}^T p_i  \cdot g(x,y, \phi_i)],
\end{aligned}
\end{equation}
where $p_i$ is the task similarity in Eq.1 in the main paper represents the probability of task $i$, equals to $P(x \ \hat{\in} \ X_{i}|\theta, \Phi)$. And as introduced in the main paper, $g(x,y,\phi_i)={\rm Softmax}(W^\top f(x,\phi_i))[y]$, represents prediction of label $y$ with task-specifc prompt for task $i$, $\phi_i$, which equals to $P(x \in X_{h,y}|x \ \hat{\in} \ X_{i}, \theta, \Phi)$.

For our method that uses aggregated prompt, the expected error ($E_1$) is as follows, as our approach eliminates the need for task prediction:
\begin{equation}
    \begin{aligned}
        E_1= &\mathbb{E}_x[{\rm -log}P(x \in X_{h,y}|\theta, \Phi)]
        \\ = &\mathbb{E}_x[{\rm -log} \ g(x,y,\sum_{i=1}^t p_i \cdot \phi_i)].
    \end{aligned}
\end{equation}
Given the assumption that the prediction of the label $y$ is concave to the combination of prompts for different tasks, according to Jensen Inequality:
\begin{equation}
    g(x,y,\sum_{i=1}^t p_i \cdot \phi_i) \ge \sum_{i=1}^t p_i  \cdot g(x,y, \phi_i).
\end{equation}
Then we have:
\begin{equation}
    \begin{aligned}
        E_1 = &\mathbb{E}_x[{\rm -log} \ g(x,y,\sum_{i=1}^t p_i \cdot \phi_i)]
        \\ \le &\mathbb{E}_x[{\rm -log} \sum_{i=1}^T p_i  \cdot g(x,y, \phi_i)] 
        \\ = &E_2.
    \end{aligned}
\end{equation}
This finishes the proof.

\end{document}